\documentclass{article}
\usepackage{spconf,amsmath,graphicx, amssymb,bbm,cleveref,booktabs,multirow,float}

\newcommand{\ms}[2]{{#1}{\footnotesize $\,\pm${#2}}}
\newcommand{\msb}[2]{{\textbf{#1}}{\footnotesize $\,\pm${#2}}}
\crefname{figure}{Fig.}{Figs.}
\crefname{table}{Table.}{Tables.}
\crefname{equation}{Eq.}{Eqs.}

\title{CLAF: Contrastive Learning with Augmented Features \\for Imbalanced Semi-Supervised Learning}
\name{Bowen Tao, Lan Li, Xin-Chun Li, De-Chuan Zhan}
\address{National Key Laboratory for Novel Software Technology, Nanjing University, China\\
School of Artificial Intelligence, Nanjing University, China }
\begin{document}
%
\maketitle
\begin{abstract}
Due to the advantages of leveraging unlabeled data and learning meaningful representations, semi-supervised learning and contrastive learning have been progressively combined to achieve better performances in popular applications with few labeled data and abundant unlabeled data. One common manner is assigning pseudo-labels to unlabeled samples and selecting positive and negative samples from pseudo-labeled samples to apply contrastive learning. However, the real-world data may be imbalanced, causing pseudo-labels to be biased toward the majority classes and further undermining the effectiveness of contrastive learning. To address the challenge, we propose Contrastive Learning with Augmented Features (CLAF). We design a class-dependent feature augmentation module to alleviate the scarcity of minority class samples in contrastive learning. For each pseudo-labeled sample, we select positive and negative samples from labeled data instead of unlabeled data to compute contrastive loss. Comprehensive experiments on imbalanced image classification datasets demonstrate the effectiveness of CLAF in the context of imbalanced semi-supervised learning. 

\end{abstract}
\section{Introduction}
\label{sec:intro}
Semi-supervised learning (SSL) has attracted much attention in recent years, owing to its potential to mitigate the demand for labeled data by leveraging unlabeled data. The primary challenge in SSL lies in learning valuable information from a large amount of unlabeled data. Representation learning empowers the capture of rich insights from labeled data, thereby reducing the difficulty of utilizing unlabeled data. Contrastive learning is an effective way to learn strong visual representations in an unsupervised manner and has been extended to supervised learning~\cite{SCL}, making it a promising approach for integration into SSL. A general pipeline of incorporating contrastive learning into SSL involves producing pseudo-labels for unlabeled data and utilizing them in a manner of pseudo-label-based contrastive learning (PCL). For a pseudo-labeled sample, PCL selects unlabeled samples sharing the same pseudo-label as positive samples and regards unlabeled samples with different pseudo-labels as negative samples. The central idea of PCL is to bring positive samples closer while pushing negative samples further apart. Through the integration of PCL, most of the existing SSL algorithms have achieved exceptional performance~\cite{CCSSL,comatch,sscl}. 
\begin{figure}[t]
    \centering
    \includegraphics[width=\linewidth]{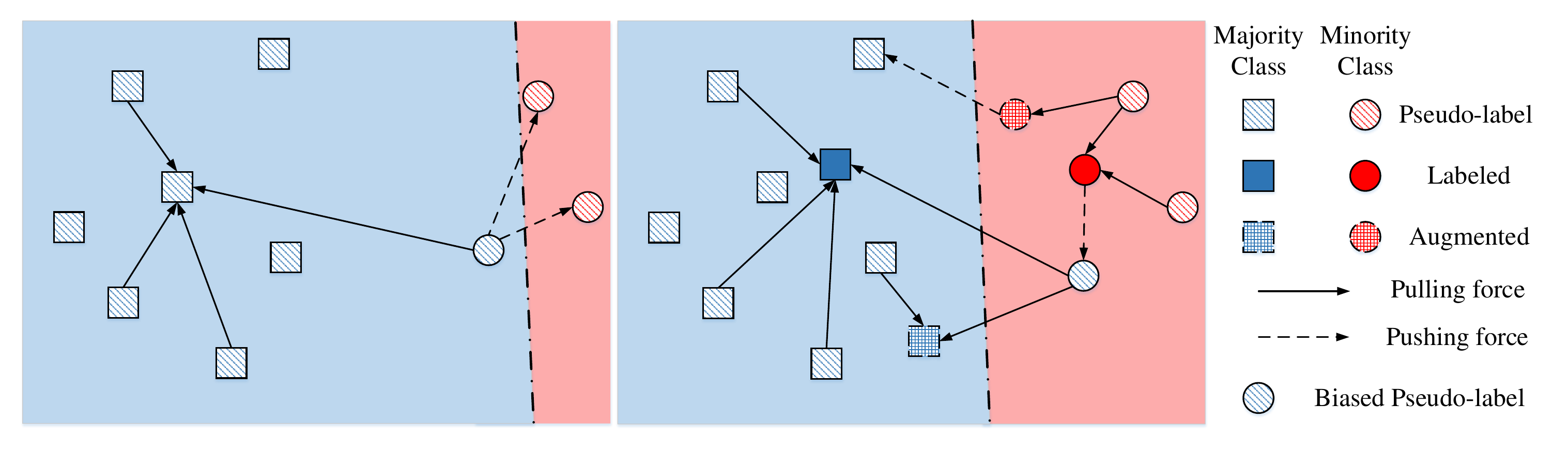}
    \caption{Illustration of PCL (left) and CLAF (right) in imbalanced SSL. Compared with PCL, CLAF adopts class-dependent feature augmentation and selects positive and negative samples from both labeled samples and augmented samples to reduce the pushing and pulling effects of samples with biased pseudo-labels on other pseudo-labeled samples.}
    \label{fig:overview}
\end{figure}

Although contrastive learning has demonstrated its efficacy in learning strong representations under SSL, these algorithms often assume class-balanced data, while many real-world data exhibit imbalanced distributions. Contrastive learning faces the risk of biased pseudo-labels and scarcity of minority class samples under imbalanced SSL. 
With class-imbalanced data, the class distribution of pseudo-labels from unlabeled data tends to exhibit towards the majority classes due to the confirmation bias~\cite{confirmationbias}.
Many pseudo-labels of majority classes are assigned to unlabeled samples that may not genuinely belong to those classes. Methods incorporating PCL tend to cluster instances with the same pseudo-labels from a specific majority class, potentially contradicting the actual relationships among unlabeled data. Additionally, the scarcity of minority class samples results in relatively poor representations of minority classes. These problems significantly constrain the representation learning capacity of contrastive learning in imbalanced SSL.
In essence, the imbalanced data distribution leads to inaccurate pseudo-labels, subsequently undermining the precision of positive and negative samples.

In this paper, we propose a method called Contrastive Learning with Augmented Features (CLAF) devised to tackle the aforementioned challenges. First, we design a class-dependent feature augmentation module to alleviate the scarcity of labeled data in minority classes. Second, in contrast to conventional PCL that exclusively selects sample pairs from unlabeled data, CLAF selects both positive and negative samples from labeled data for each pseudo-labeled sample to reduce the influence of biased pseudo-labels as shown in~\cref{fig:overview}.

\section{Related Works}
\noindent \textbf{Semi-supervised learning (SSL)}: SSL learns from labeled data in conjunction with a large number of unlabeled data. Pseudo-labeling is a widely used SSL method, which uses the model's predictions to label data and retrains the model with the artificial labels~\cite{confirmationbias}. FixMatch~\cite{fixmatch} integrates consistency regularization and pseudo-labeling to align the predictions between weakly and strongly augmented unlabeled images.

\noindent \textbf{Contrastive learning under SSL}: Previous contrastive-based SSL works are almost two-stage ones. SelfMatch~\cite{selfmatch} adopts contrastive learning to pre-train a backbone and then fine-tune it based on augmentation consistency regularization. Existing SSL methods that build upon FixMatch mostly utilize pseudo-labels for contrastive learning~\cite{comatch}. To make use of the features learned by different loss functions and class-specific priors, SsCL~\cite{sscl} adopts the pseudo-labeling strategy with cross-entropy loss and instance discrimination with contrastive loss, jointly optimizing the two losses with a shared backbone in an end-to-end way. To address the confirmation bias due to the noise contained in pseudo-labels, CCSSL~\cite{CCSSL} introduces a class-aware contrastive module and focuses learning on unlabeled samples with pseudo-labels. 

\section{Preliminary}
\subsection{Problem Setup} For a $K$-class semi-supervised image classification task, we are given labeled data $\mathcal{X}=\{ (x_{n}, y_{n}) \}^{N}_{n=1}$ and unlabeled data $\mathcal{U}=\{u_{m} \}^{M}_{m=1}$ to train a model $f$ comprising a feature encoder $f^{\text{enc}}_\theta$ followed by a linear classifier $f_\phi^\text{cls}$, where $\theta$ and $\phi$ correspond to the parameters of $f^{\text{enc}}_\theta$ and $f_\phi^\text{cls}$ respectively. 
For labeled data, the prediction $f(x)$ of a image $x$ is learned from $\mathcal{L}_{\text{cls}}$ (\textit{e.g.}, cross-entropy) and its label $y$.
For unlabeled data, a pseudo-label $\hat{p} \in \mathbb{R}^K$ is utilized in unsupervised loss $\mathcal{L}_{u}=\Phi_{u} \left(\hat{p}, f(u) \right)$, where $\Phi_{u}$ can be implemented via entropy~\cite{entropy} or consistency regularization~\cite{meanteachers}, depending on the SSL methods adopted.

Take FixMatch~\cite{fixmatch} as an example, the pseudo-label $\hat{p}=\textnormal{OneHot}\left(\mathop{\mathrm{argmax}}_k p^{{(w)}}_k \right)$ with $p^{{(w)}}=f\left(\mathcal{A}_w(u) \right)$ provides the target for the prediction $p^{{(s)}}=f\left(\mathcal{A}_s(u) \right)$ with some confident ones to the cross-entropy loss $\mathcal{H}$ as follow: 
\begin{equation}
    \label{eqn:fixmatch}
    \Phi_u(\hat{p},\,p^{{(s)}}) = \mathbbm{1}\left(\max_k p^{{(w)}}_k \geq \tau\right)\,\mathcal{H} \left(\hat{p},\,p^{{(s)}} \right), 
\end{equation}
where $\mathcal{A}_w$ and $\mathcal{A}_s$ correspond to weak augmentation and strong augmentation~\cite{randaugment} respectively. 
\subsection{DASO}
DASO~\cite{DASO} is a comprehensive framework for imbalanced SSL incorporating distribution-aware blending for both linear and semantic pseudo-labels. The linear and semantic pseudo-label, $\hat{p}$ and $\hat{q}$ are generated by passing $z^{(w)}=f^{\text{enc}}_\theta(\mathcal{A}_{w}(u))$ through linear and similarity-based classifier respectively. Subsequently, the final pseudo-label $\hat{p}^{\prime}$ is derived through the fusion of $\hat{p}$ and $\hat{q}$ and serves as the target in $\mathcal{L}_u=\Phi_u(\hat{p}^{\prime},\,p)$.

The linear pseudo-label $\hat{p}$ is obtained by applying the softmax function to the output of the linear classifier: $\hat{p} = \sigma(f^{\text{cls}}_{\phi}(z^{(w)}))$. The semantic pseudo-label $\hat{q}$ is derived from a similarity-based classifier. Specifically, DASO constructs a set of class prototypes $\mathbf{C}  = \{c_{k} \}^{K}_{k=1}$ from $\mathcal{X}$ and a queue $\mathbf{Q} =\{Q_{k}\}^{K}_{k=1}$ where $Q_{k}$ denotes a feature queue for class $k$ with a fixed size $|Q_k|$. The class prototype $c_k$ for each class $k$ can be obtained simply by averaging the feature points in the feature queue $Q_{k}$. DASO measures the per-class similarity between a feature point and class prototypes:
\begin{equation}
    \label{eq:sem_pl}
    q = \sigma(\text{sim}(z^{(w)}, \mathbf{C})/T_{proto}),
\end{equation}
where $\text{sim}(\cdot, \cdot)$ represents cosine similarity and $T_{proto}$ is a temperature hyper-parameter. To prevent an imbalanced prototype representation arising from class-imbalanced labeled data, DASO fixes the size of $Q_k$ for all classes to the same amount, which can compensate for the prototypes of the minority classes with earlier samples remaining in the queue. To stabilize the movement of class prototypes in feature space during training, DASO employs a momentum encoder $f^{\text{enc}}_{\theta^{\prime}}$ with the same architecture as $f^{\text{enc}}_{\theta}$, where $\theta'$ is the exponential moving average (EMA) of $\theta$ with momentum ratio $\rho$: $\theta^{\prime} \leftarrow \rho \theta^{\prime} + (1-\rho)\theta$.
\section{Method}
\subsection{Class-dependent Feature Augmentation}
DASO introduces a balanced queue to ensure equilibrium between minority and majority class samples. Notably, a significant portion of minority class features in the queue is generated from the same labeled data. To enhance data diversity and alleviate the scarcity of labeled data in minority classes, we employ feature augmentation (FA) within a batch to increase the count of labeled features for minority classes by blending unlabeled data features with labeled data features while preserving the label of the original labeled sample, which is inspired by~\cite{cossl,mixup,imbalance}. The augmented feature is generated as:
\begin{equation}
    z_{aug} = \lambda z^{\prime(l)} + (1 - \lambda) z^{\prime(w)},
\end{equation}
where $z^{\prime(l)} = f^{\text{enc}}_{\theta'}(x)$ and $z^{\prime(w)} = f^{\text{enc}}_{\theta'}(\mathcal{A}_w(u))$. $\lambda$ is the mixture coefficient sampled from a Beta distribution denoted as $\text{Beta}(\alpha, \alpha)$. To ensure the validity of the label for the augmented feature, we consider $\lambda$ with a value at least $\mu$: $\lambda = \max(\lambda, 1 - \lambda, \mu)$. The FA is applied with a probability that depends on the count of labeled data for each class. Consequently, the more labeled data a class has, the less augmented feature is synthesized. Formally, given a labeled sample from class $k$, we apply FA with probability $P_k$ defined as:
\begin{equation}
\label{eq:mixprob}
    P_k = \frac{N_1 - N_k}{N_1},
\end{equation}
where $N_k$ is the number of samples of class $k$ and $N_1$ is the number of samples of the class with the most labeled data. The class-dependent probability encourages more augmented features for minority classes. 

We perform concurrent updates of $Q_k$ for all classes by pushing new labeled features and augmented features within the batch and removing the oldest ones when $Q_k$ is full.
\subsection{Contrastive Learning with Augmented Features}
To reduce the impact of biased pseudo-labels and utilize unlabeled data, we apply contrastive learning using both unlabeled and labeled data. For an unlabeled sample with a pseudo-label, we bring it close to labeled samples sharing the same label as the pseudo-label and push it away from labeled samples with different labels from the pseudo-label.

Following the common approaches in contrastive learning~\cite{simclr}, we adopt the encoder-projection head structure in our method. Both raw feature and augmented feature are passed through the projection head to obtain corresponding embedding $e$. We construct an extra embedding queue $\mathbf{E}$ to store embeddings for features with labels, which is updated simultaneously with the feature queue $\mathbf{Q}$. For unlabeled samples, we establish a confidence vector $\mathbf{s}$ based on the confidence scores of the model's predictions. Each element $s_{i}$ in $\mathbf{s}$ is defined as:
\begin{equation}
        s_{i} = \begin{cases}
\max(\hat{p}^{\prime}_{i}), & \text{if } \max(\hat{p}^{\prime}_{i}) > \tau, \\
0, & \text{otherwise}.
\end{cases}
\label{eq:conf}
\end{equation}
where $i$ is the index of the unlabeled sample. Given the presence of embeddings from augmented features, we construct a label confidence vector $\mathbf{v}$ based on the mixture coefficient:
\begin{equation}
    v_{i} = \begin{cases}
        \lambda_i, & \text{if } e_{i} \text{ corresponds to an augmented feature}, \\
        1, & \text{otherwise}.
    \end{cases}
    \label{eq:mixcof}
\end{equation}
where $i$ is the index of embedding in the embedding queue. To measure the weights for positive pairs in contrastive loss function, we obtain a weight matrix $W$ by multiplying elements of $\mathbf{s}$ and $\mathbf{v}$. Each element $w_{ij}$ in $W$ is defined as $w_{ij} = s_i \cdot v_j$, where $i$ and $j$ represent the indices of unlabeled samples in a batch and embeddings in the embedding queue of the pseudo-label class. The contrastive loss $\mathcal{L}_c$ can be defined as:
\begin{equation}
    \mathcal{L}_c = \frac{1}{B}\sum_{i=1}^B\mathcal{L}_{c,i},
\end{equation}
where $B$ is the batch size of unlabeled samples. $\mathcal{L}_{c,i}$ has the following format:
\begin{equation}
    \mathcal{L}_{c,i} = - \frac{1}{|E_{p_i}|}\sum_{p=1}^{|E_{p_i}|}w_{ip} \cdot \log\frac{\exp(\text{sim}(e^{(s)}_i, e^{(l)}_p)/t)}{\sum_{k=1}^K\sum_{j=1}^{|E_k|}\exp(\text{sim}(e^{(s)}_i,e^{(l)}_j)/t)},
\end{equation}
where $E_{p_i}$ denotes the embedding queue of the pseudo-label class and $|E_{p_i}|$ represents the capacity of $E_{p_i}$. $e^{(s)}_i$ and $e^{(l)}_p$ are embeddings from $z_i^{(s)}=f^{\text{enc}}_{\theta}(\mathcal{A}_s(u_i))$ and $E_{p_i}$ respectively. $t$ is the temperature hyper-parameter. We calculate total loss using a weighted sum of supervised loss $\mathcal{L}_{cls}$, semi-supervised loss $\mathcal{L}_u$, semantic alignment loss $\mathcal{L}_{align}$ and contrastive loss $\mathcal{L}_{c}$. The final CLAF objective is as below:
\begin{equation}
    \mathcal{L}_{CLAF} = \mathcal{L}_{cls} + \lambda_u\mathcal{L}_u + \lambda_{align}\mathcal{L}_{align} + \lambda_c\mathcal{L}_{c},
    \label{eq:mixdasoloss}
\end{equation}
where both $\mathcal{L}_{cls}$ and $\mathcal{L}_{u}$ with $\lambda_u$ come from the base SSL learner, and $\mathcal{L}_{align}$ is introduced from DASO. $\lambda_c$ is the weight for contrastive loss.
\section{Experiments}
\begin{table*}[ht]
    \centering
    \scalebox{0.75}{
    \begin{tabular}{lcccccccc}\toprule
         &  \multicolumn{4}{c}{CIFAR10-LT} & \multicolumn{4}{c}{CIFAR100-LT}\\
         & \multicolumn{2}{c}{$\gamma=100$} & \multicolumn{2}{c}{$\gamma=150$} & \multicolumn{2}{c}{$\gamma=10$} & \multicolumn{2}{c}{$\gamma=20$}\\
         \cmidrule(lr){2-3} \cmidrule(lr){4-5} \cmidrule(lr){6-7} \cmidrule(l){8-9}
         \multirow{2}{*}{Algorithms} & $N_1=500$ & $N_1=1500$ & $N_1=500$ & $N_1=1500$ & $N_1=50$ & $N_1=150$ & $N_1=50$ & $N_1=150$ \\
         & $M_1=4000$ & $M_1=3000$ & $M_1=4000$ & $M_1=3000$ & $M_1=400$ & $M_1=300$ & $M_1=400$ & $M_1=300$ \\
         \cmidrule(r){1-1} \cmidrule(lr){2-3} \cmidrule(lr){4-5} \cmidrule(lr){6-7} \cmidrule(l){8-9}
         Supervised$^*$ & \ms{47.3}{0.95} & \ms{61.9}{0.41} & \ms{44.2}{0.33} & \ms{58.2}{0.29} & \ms{29.6}{0.57} & \ms{46.9}{0.22} & \ms{25.1}{1.14} & \ms{41.2}{0.15} \\
         ~~w/ LA$^*$~\cite{LA} & \ms{53.3}{0.44} & \ms{70.6}{0.21} & \ms{49.5}{0.40} & \ms{67.1}{0.78} & \ms{30.2}{0.44} & \ms{48.7}{0.89} & \ms{26.5}{1.31} & \ms{44.1}{0.42}\\  \cmidrule(r){1-1} \cmidrule(lr){2-3} \cmidrule(lr){4-5} \cmidrule(lr){6-7} \cmidrule(l){8-9}

         FixMatch$^*$~\cite{fixmatch} & \ms{67.8}{1.13} & \ms{77.5}{1.32} & \ms{62.9}{0.36}  & \ms{72.4}{1.03}  & \ms{45.2}{0.55} & \ms{56.5}{0.06} & \ms{40.0}{0.96} & \ms{50.7}{0.25} \\
        ~~w/ DARP$^*$~\cite{DARP} & \ms{74.5}{0.78} & \ms{77.8}{0.63}  & \ms{67.2}{0.32} & \ms{73.6}{0.73}  & \ms{49.4}{0.20} & \ms{58.1}{0.44} & \ms{43.4}{0.87}  & \ms{52.2}{0.66} \\
        ~~w/ CReST+$^*$~\cite{CREST} & \ms{76.3}{0.86} & \ms{78.1}{0.42} & \ms{67.5}{0.45} & \ms{73.7}{0.34} & \ms{44.5}{0.94} & \ms{57.4}{0.18} & \ms{40.1}{1.28} & \ms{52.1}{0.21} \\
        ~~w/ DASO$^*$~\cite{DASO} & \ms{76.0}{0.37} & \ms{79.1}{0.75}   & \ms{70.1}{1.81} & \ms{75.1}{0.77}  & \ms{49.8}{0.24} & \ms{59.2}{0.35} & \ms{43.6}{0.09} & \ms{52.9}{0.42} \\
        ~~w/ CLAF (Ours) & \msb{76.4}{0.46} & \msb{80.6}{0.65} & \msb{72.0}{0.74} & \msb{75.9}{0.29} & \msb{50.9}{0.11} & \msb{59.8}{0.29} & \msb{44.5}{0.83} & \msb{54.1}{0.28} \\ 
        \cmidrule(r){1-1} \cmidrule(lr){2-3} \cmidrule(lr){4-5} \cmidrule(lr){6-7} \cmidrule(l){8-9}

        FixMatch$+$LA$^*$~\cite{LA} & \ms{75.3}{2.45} & \ms{82.0}{0.36} & \ms{67.0}{2.49} & \ms{78.0}{0.91} & \ms{47.3}{0.42} & \ms{58.6}{0.36} & \ms{41.4}{0.93} & \ms{53.4}{0.32} \\
        ~~w/ DASO$^*$~\cite{DASO} & \ms{77.9}{0.88} & \ms{82.5}{0.08} & \ms{70.1}{1.68} & \ms{79.0}{2.23} & \ms{50.7}{0.51} & \ms{60.6}{0.71} & \ms{44.1}{0.61} & \ms{55.1}{0.72} \\
        ~~w/ CLAF (Ours) & \msb{78.8}{0.59} & \msb{83.1}{0.32} & \msb{72.8}{1.39} & \msb{79.3}{0.33} & \msb{51.1}{0.25} & \msb{60.9}{0.22} & \msb{46.1}{0.19} & \msb{55.6}{0.51} \\
        \bottomrule
    \end{tabular}
    }
    \caption{Comparison of accuracy(\%) with different methods on CIFAR10-LT and CIFAR100-LT under various $\gamma$. $^*$: reported by~\cite{DASO}. We mark the best results as bold. Our method CLAF consistently outperforms all the baselines under diverse settings.}
    \label{tab:claf}
\end{table*}
\subsection{Experimental Setup}
\subsubsection{Datasets}
Following common practice~\cite{DASO}, we create CIFAR10-LT and CIFAR100-LT for imbalanced SSL by exponentially decreasing the count of images from the head class to the tail class. We denote the head class size as $N_1(M_1)$ and the imbalance ratio as $\gamma$. We set $N_k = N_1\cdot \gamma^{-\frac{k-1}{K-1}}$ for labeled data and $M_k=M_1\cdot\gamma^{-\frac{k-1}{K-1}}$ for unlabeled data. For common settings~\cite{DASO}, we set $N_1=500$, $M_1=4000$ and $N_1=1500$, $M_1=3000$ for CIFAR10-LT, and $N_1=50$, $M_1=400$ and $N_1=150$, $M_1=300$ for CIFAR100-LT. We report results of imbalance ratio $\gamma=100$ and $150$ for CIFAR10-LT and $\gamma=10$ and $20$ for CIFAR100-LT. 
\subsubsection{Training and evaluation}
We conduct experiments under the same codebase with DASO~\cite{DASO} for fair comparison. We adopt Wide ResNet-28-2~\cite{WRN} as our backbone on CIFAR10-LT and CIFAR100-LT. We apply FA in the last 20\% iterations and set $\mu$ to 0.8 to meet the requirements of FA for structured representation space. $\lambda_c$ and $t$ are set to 1.0 and 0.07 for all experiments. All hyper-parameters and training details follow DASO~\cite{DASO}. For evaluation, we use the EMA network with parameters updating every training step~\cite{DASO}. We measure the top-1 accuracy on test images every 500 iterations and report the median of the accuracy of the last 20 evaluations. We report the mean and standard deviation of three independent runs.

\subsection{Results on CIFAR10/CIFAR100-LT}
We report the results of CLAF on CIFAR10-LT and CIFAR100-LT under various settings in~\cref{tab:claf}. We compare CLAF with DARP~\cite{DARP}, CReST+~\cite{CREST} and DASO~\cite{DASO} on FixMatch. The results indicate CLAF achieves superior accuracy compared with baselines on different benchmarks. The results of different methods on re-balancing FixMatch via LA~\cite{LA} show CLAF can benefit from debiasing pseudo-labels. It is noticeable that CLAF always exhibits performance improvements over DASO in all cases, which verifies the effectiveness of CLAF in representation learning under imbalanced SSL.
\subsection{Ablation Study}
We perform ablation studies on CIFAR10-LT and investigate the impact of FA. We report the results of CLAF and CLAF without FA in~\cref{tab:ablation}. As previously discussed, FA mainly contributes to augmenting features for minority classes and providing minority class features for contrastive learning. The performance gap between CLAF and CLAF without FA indicates that naive contrastive learning brings marginal improvements and FA is beneficial for contrastive learning in imbalanced SSL. 
\begin{table}[ht]
    \centering
    \scalebox{0.75}{
    \begin{tabular}{lcccc}\toprule
         & \multicolumn{2}{c}{$\gamma=100$} & \multicolumn{2}{c}{$\gamma=150$} \\
         \cmidrule(lr){2-3} \cmidrule(lr){4-5} 
         \multirow{2}{*}{Algorithm} & $N_1=500$ & $N_1=1500$ & $N_1=500$ & $N_1=1500$ \\
         & $M_1=4000$ & $M_1=3000$ & $M_1=4000$ & $M_1=3000$ \\
         \cmidrule(r){1-1} \cmidrule(lr){2-3} \cmidrule(lr){4-5}
         CLAF & \msb{76.4}{0.46} & \msb{80.6}{0.65} & \msb{72.0}{0.74} & \msb{75.9}{0.29} \\
         ~~w/o FA & \ms{76.1}{0.25} & \ms{79.9}{0.24} & \ms{70.8}{2.15} & \ms{75.5}{0.41} \\ \bottomrule
    \end{tabular}
    }
    \caption{Ablation study of FA in CLAF on CIFAR10-LT.}
    \label{tab:ablation}
\end{table}
\subsection{Analysis}
\begin{figure}
    \centering
    \scalebox{0.9}{
    \includegraphics[width=\linewidth]{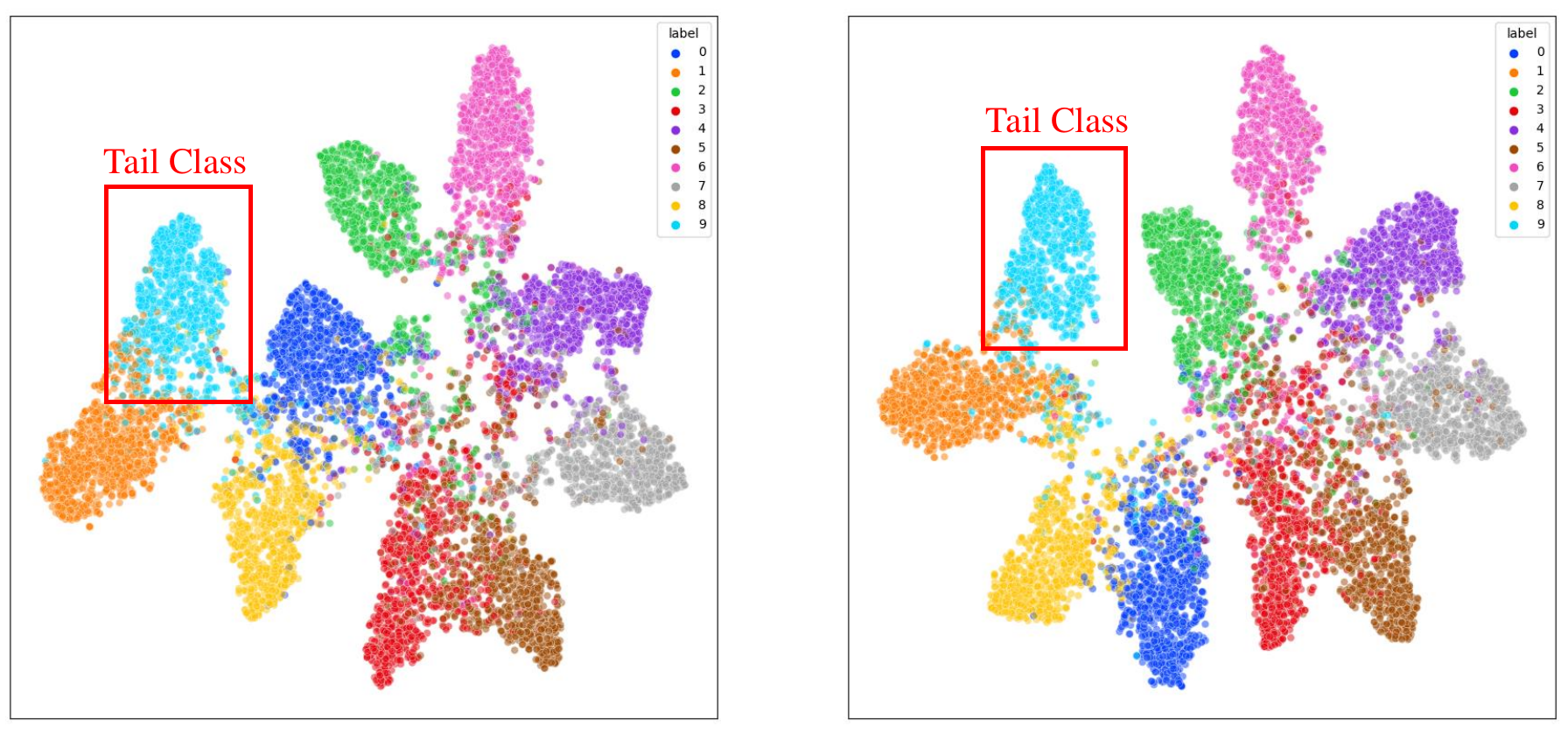}
    }
    \caption{Feature space visualization of CIFAR10-LT test data using DASO (left) and CLAF (right). CLAF has clear decision boundaries for tail class in feature space.}
    \label{fig:tsne}
\end{figure}

To assess the representation learning capacity of contrastive learning, we present t-SNE~\cite{tsne} visualization of CIFAR10-LT test data features obtained from DASO and CLAF. As shown in~\cref{fig:tsne}, tail class features in CLAF exhibit distinct decision boundaries while they are close to majority class features in DASO. CLAF achieves the accuracy of $65.3\%$ for the 3-least common classes, which is better than $60.8\%$ in DASO. The results suggest that CLAF has superior representations for minority classes compared to DASO.

\section{Conclusion}
We propose Contrastive Learning with Augmented Features (CLAF) to apply contrastive learning in imbalanced SSL. We design a class-dependent feature augmentation module to alleviate the scarcity of minority class samples. In contrast to conventional PCL, we select positive and negative samples from labeled data to reduce the impact of biased pseudo-labels. Our experimental results demonstrate that CLAF outperforms the baselines on imbalanced image datasets under various settings, confirming that CLAF exhibits a remarkable capacity for representation learning in imbalanced SSL.




\bibliographystyle{IEEEbib}
\bibliography{CLAF}

\end{document}